\documentclass[10pt,twocolumn,letterpaper]{article}

\usepackage[pagenumbers]{cvpr} 
\usepackage[table,dvipsnames]{xcolor}
\usepackage{color}
\usepackage{multirow}
\usepackage[dvipsnames]{xcolor}

\usepackage[misc]{ifsym}

\usepackage{amsmath,amsfonts,bm}

\def\eqref#1{equation~\ref{#1}}

\def\1{\bm{1}}

\def\mA{{\bm{A}}}
\def\mB{{\bm{B}}}

\def\mD{{\bm{D}}}

\def\mG{{\bm{G}}}

\def\mI{{\bm{I}}}

\def\mK{{\bm{K}}}

\def\mM{{\bm{M}}}

\def\mP{{\bm{P}}}
\def\mQ{{\bm{Q}}}

\def\mW{{\bm{W}}}

\def\mY{{\bm{Y}}}

\DeclareMathAlphabet{\mathsfit}{\encodingdefault}{\sfdefault}{m}{sl}
\SetMathAlphabet{\mathsfit}{bold}{\encodingdefault}{\sfdefault}{bx}{n}

\def\sR{{\mathbb{R}}}

\definecolor{cvprblue}{rgb}{0.21,0.49,0.74}
\usepackage[pagebackref,breaklinks,colorlinks,citecolor=cvprblue]{hyperref}
\definecolor{aliceblue}{rgb}{0.94, 0.97, 1.0}
\definecolor{darkgreen}{HTML}{539165}

\makeatletter
\newcommand{\thickhline}{%
 \noalign {\ifnum 0=`}\fi \hrule height 1pt
 \futurelet \reserved@a \@xhline
}
\makeatother
\newcommand{\increase}[1]{{
  \fontsize{7.5pt}{0.5em}\selectfont({\color{darkgreen}{$\uparrow$~\textbf{#1}}})
}}
\newcommand{\decrease}[1]{{
  \fontsize{7.5pt}{0.5em}\selectfont({\color{purple}{$\downarrow$~\textbf{#1}}})
}}
\renewcommand{\thefootnote}{}

\DeclareMathAlphabet\mathbfcal{OMS}{cmsy}{b}{n}

\title{GraCo: Granularity-Controllable Interactive Segmentation}

\author{Yian Zhao\textsuperscript{1,3} \quad Kehan Li\textsuperscript{1,3} \quad Zesen Cheng\textsuperscript{1,3} \quad Pengchong Qiao\textsuperscript{1,2,3} \quad Xiawu Zheng\textsuperscript{4} \quad \\
Rongrong Ji\textsuperscript{4} \quad Chang Liu\textsuperscript{5} \quad Li Yuan\textsuperscript{1,2,3} \quad Jie Chen\textsuperscript{1,2,3\ \Letter} \and
\small\textsuperscript{1}School of Electronic and Computer Engineering, Peking University, Shenzhen, China
 \quad
\small\textsuperscript{2}Peng Cheng Laboratory, Shenzhen, China \\
\small\textsuperscript{3}AI for Science (AI4S)-Preferred Program, Peking University Shenzhen Graduate School, China \\
\small\textsuperscript{4}Key Laboratory of Multimedia Trusted Perception and Efficient Computing, Ministry of Education of China, Xiamen University, China \\
\small\textsuperscript{5}Department of Automation and BNRist, Tsinghua University, Beijing, China \\
\small \href{mailto:zhaoyian@stu.pku.edu.cn}{zhaoyian@stu.pku.edu.cn}  \quad \small \href{mailto:jiechen2019@pku.edu.cn}{jiechen2019@pku.edu.cn}
\vspace*{-5mm}
}
\begin{document}
\maketitle
\begin{abstract}
Interactive Segmentation~(IS) segments specific objects or parts in the image according to user input.
Current IS pipelines fall into two categories: single-granularity output and multi-granularity output. The latter aims to alleviate the spatial ambiguity present in the former.
However, the multi-granularity output pipeline suffers from limited interaction flexibility and produces redundant results.
In this work, we introduce \textbf{Gra}nularity-\textbf{Co}ntrollable Interactive Segmentation~(\textbf{GraCo}), a novel approach that allows precise control of prediction granularity by introducing additional parameters to input. 
This enhances the customization of the interactive system and eliminates redundancy while resolving ambiguity.
Nevertheless, the exorbitant cost of annotating multi-granularity masks and the lack of available datasets with granularity annotations make it difficult for models to acquire the necessary guidance to control output granularity.
To address this problem, we design an any-granularity mask generator that exploits the semantic property of the pre-trained IS model to automatically generate abundant mask-granularity pairs without requiring additional manual annotation. 
Based on these pairs, we propose a granularity-controllable learning strategy that efficiently imparts the granularity controllability to the IS model.
Extensive experiments on intricate scenarios at object and part levels demonstrate that our GraCo has significant advantages over previous methods.
This highlights the potential of GraCo to be a flexible annotation tool, capable of adapting to diverse segmentation scenarios. The project page: \href{https://zhao-yian.github.io/GraCo}{https://zhao-yian.github.io/GraCo}.
\end{abstract}
\footnote{\Letter\ Corresponding author.}
\vspace*{-4mm}
\section{Introduction}
\label{sec:intro}

Interactive Segmentation~(IS) aims to segment specific objects or parts according to user interactions, providing a pixel-level interactive AI system that follows human intent.
Recently, remarkable progress has been achieved in IS, resulting in various applications such as controllable image generation~\cite{rombach2022high,zhang2023adding}, image editing~\cite{brooks2023instructpix2pix,kawar2023imagic}, and the well-known pixel-level annotation.
Extensive research has been undertaken on various types of interactive information, such as bounding boxes~\cite{lempitsky2009image,xu2017deep}, scribbles~\cite{li2004lazy,grady2006random,bai2014error}, and clicks~\cite{xu2016deep,jang2019interactive,lin2020interactive,sofiiuk2020f,chen2021conditional,lin2022focuscut,chen2022focalclick,liu2023simpleclick}.
Among them, the click-based interaction becomes mainstream due to its simplicity and well-established training and evaluation protocols.

\begin{figure}[!t]
\centering
\includegraphics[width=\linewidth]{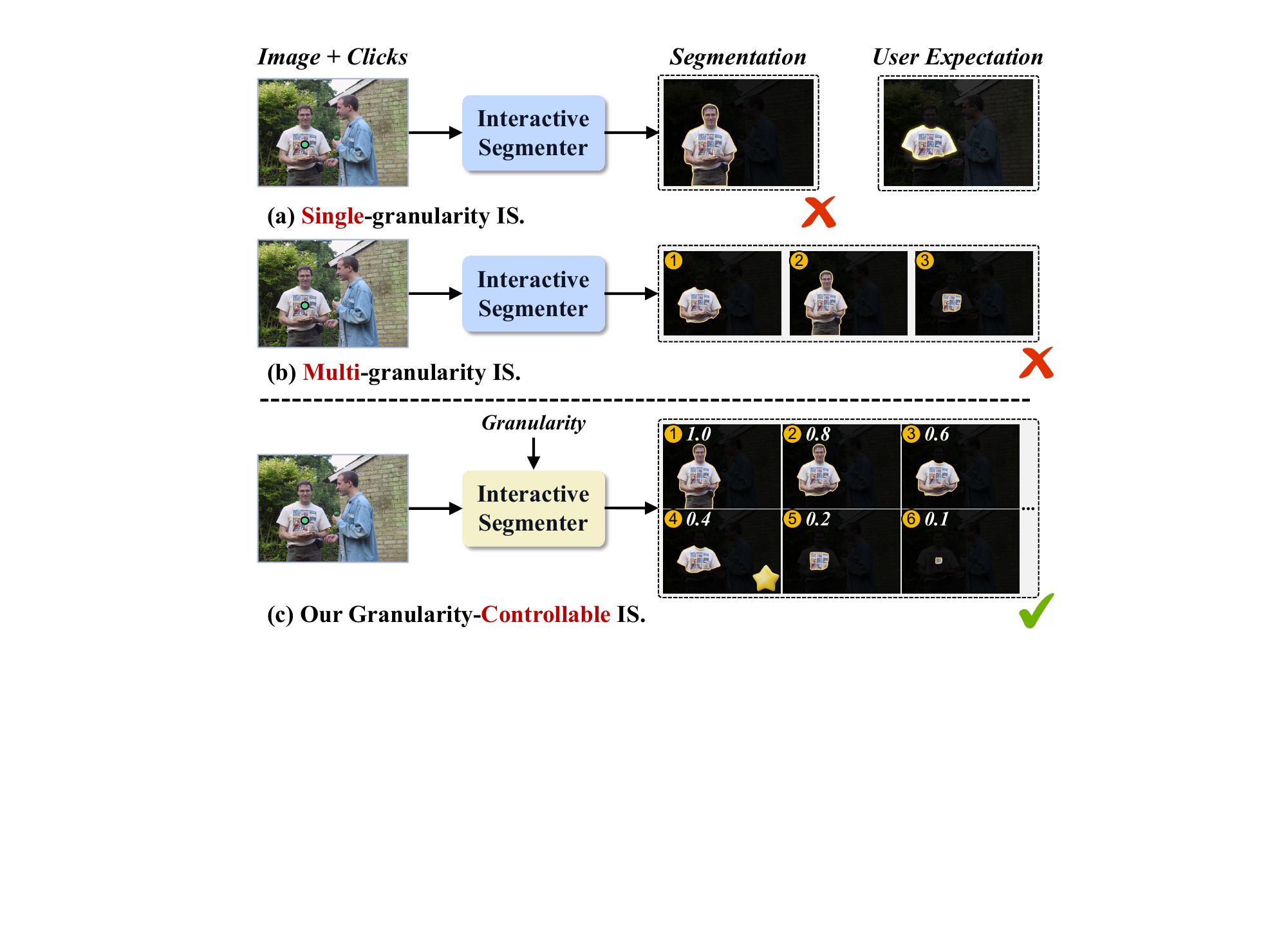} 
\vspace*{-5mm}
\caption{
\textbf{(a)}: Single-granularity IS ignores spatial ambiguity.
\textbf{(b)}: Multi-granularity IS is limited in the number of outputs and produces redundant results.
\textbf{(c)}: Our Granularity-Controllable IS allows precise control of output granularity to match user expectations by attaching additional parameters to the input. 
}
\vspace*{-7mm}
\label{fig:motivation}
\end{figure}
The current click-based IS methods are based on deep learning technology. Xu \etal~\cite{xu2016deep} first introduces this technology to formulate IS and establishes training and evaluation protocols.
Specifically, clicks are typically encoded into distance maps and then combined with the image to send the semantic segmentation model for interactive association training between clicks and GT masks.
The emergence of SAM~\cite{kirillov2023segment} strengthens the advancement of IS and proposes multi-granularity output pipelines to alleviate spatial ambiguity.
The ambiguity refers to the concept that, given an interaction click, the desired segmentation region for the user may be the concept of objects with different parts nearby.
However, this multi-granularity output pipeline suffers from limited scalability and produces redundant results, requiring the selection of the optimal mask based on confidence or user expectations.

Intuitively, the spatial ambiguity arises from the sparse clicks information supplied by the user, which fails to impose sufficient constraints for the model to establish a distinctive dense mask.
To address this, we aim to achieve \textbf{Gra}nularity-\textbf{Co}ntrollable Interactive Segmentation~(\textbf{GraCo}), which introduces a granularity control parameter to the input to explicitly constrain the model. For instance, the granularity can be controlled by a value ranging from 0 to 1, where a lower value corresponds to a finer granularity and vice versa, as shown in \Cref{fig:motivation}.
This approach allows precise control of prediction granularity, thereby enhancing the customization of pixel-level AI systems for human-machine interaction and eliminating redundancy while resolving ambiguity. 
However, the exorbitant cost of annotating multi-granularity masks and the lack of available datasets with granularity annotations corresponding to the masks make it difficult for models to acquire the necessary guidance to control output granularity.

To acquire the any-granularity masks and granularity annotations at a low cost, we design an Any-Granularity mask Generator~({AGG}) that is fully automated and does not require any additional manual annotation.
Specifically, AGG consists of two key components: a mask engine and a granularity estimator.
For the mask engine, we observe that object-level pre-trained IS models~(\eg, SimpleClick~\cite{liu2023simpleclick}) demonstrate the semantic property in delineating local concepts and object parts via appropriate interaction signals, which has the potential to generate proposals of any granularity, shape and intricacy.
Based on this observation, we propose the multi-granularity loop simulation to automatically simulate the human-in-the-loop mechanism and generate diverse interaction signals to drive the mask engine.
To estimate the granularity of the masks, we design the granularity estimator and establish computational rules from both the scale and semantic perspectives to ensure that the model behaviour is consistent with human cognition.
Based on the mask-granularity pairs generated by AGG, we develop a simple yet efficient granularity-controllable learning~(GCL) strategy, which incorporates the granularity embedding into the input and employ LoRA~\cite{hu2021lora} technology. 
This enables the IS model to efficiently possesses granularity controllability while maintaining the original IS performance without requiring extensive computational cost.

%
To evaluate the performance of the IS models in multi-granularity scenarios, we follow standard protocols~\cite{xu2016deep} and conduct experiments on both object and part level benchmarks.
For the object-level, we perform evaluation on four commonly used datasets including GrabCut~\cite{rother2004grabcut}, Berkeley~\cite{mcguinness2010comparative}, SBD~\cite{hariharan2011semantic}, and DAVIS~\cite{perazzi2016benchmark}.
For the part-level, we employ the part segmentation datasets PascalPart~\cite{chen2014detect} and PartImageNet~\cite{he2022partimagenet}.
Thanks to the abundant mask-granularity pairs generated by AGG and the GCL strategy, the pre-trained IS model efficiently grasps the granularity controllability, achieving inspiring performance across all benchmarks on both levels.
Specifically, our GraCo surpasses the state-of-the-art single-granularity IS methods on all benchmarks, especially on part-level benchmarks. Furthermore, GraCo outperforms the multi-granularity IS approach SAM~\cite{kirillov2023segment} on all benchmarks and achieves comparable performance on SA-1B~\cite{kirillov2023segment}.

The main contributions can be summarized as:
(\romannumeral1). We propose granularity-controllable interactive segmentation, which allows precise control of prediction granularity, thereby enhancing the flexibility of IS models and eliminating redundancy while resolving ambiguity;
(\romannumeral2). We explicitly exploit the semantic property of the pre-trained IS models and design a fully automated any-granularity mask generator to generate abundant mask-granularity pairs;
(\romannumeral3). We propose granularity-controllable learning strategy that enables the IS model to achieve inspiring performance on all benchmarks at both object and part levels.

\section{Related Work}
\label{sec:related}
\begin{figure*}[t]
    \centering
    \includegraphics[width=\textwidth]{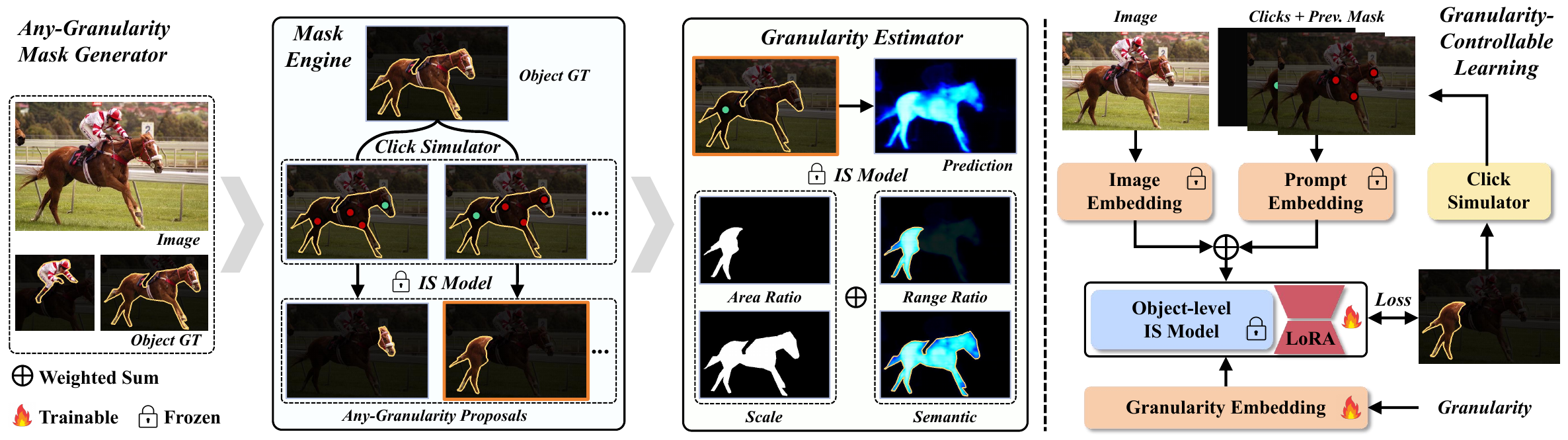}
    \vspace*{-5mm}
    \caption{
        \textbf{Illustration of our granularity-controllable interactive segmentation.}
        Our GraCo consists of two stages. For the first stage, the Any-Granularity mask Generator~(AGG) is designed to automatically generate any-granularity proposals~(mask engine) and granularity annotations~(granularity estimator) based on the object GT, without requiring additional manual annotation.
        For the second stage, the mask-granularity pairs generated by AGG are utilized to perform Granularity-Controllable Learning~(GCL) on the object-level pre-trained IS model, enabling the model to efficiently possesses granularity controllability.
    }
    \vspace*{-5mm}
    \label{fig:overview}
\end{figure*}
\noindent \textbf{Single-granularity Interactive Segmentation.}
Interactive Segmentation~(IS) is a thriving field due to its adaptability and broad applications.
Early studies for IS typically utilize the low-level features and build optimization-based methods, including graph cut with max-flow algorithm~\cite{boykov2004experimental}, random walk~\cite{grady2006random}, geodesic distance~\cite{bai2007geodesic}, and star-convexity~\cite{gulshan2010geodesic}.
These methods usually suffer from unsatisfactory performance when processing complex surroundings.
DIOS~\cite{xu2016deep} first introduces deep learning for IS, which proposes a click sampling strategy and establishes training and evaluation protocols.
Based on this framework, researchers propose a range of optimization schemes from the perspectives of global segmentation and local refinement.
FCA-Net~\cite{lin2020interactive} highlights the significance of first click. RITM~\cite{sofiiuk2022reviving} propose an iterative sampling strategy in training. BRS~\cite{jang2019interactive,sofiiuk2020f} introduces online optimization to correct mislabeled pixels.
CDNet~\cite{chen2021conditional} designs a conditional diffusion module to optimize segmentation. FocusCut~\cite{lin2022focuscut} and FocalClick~\cite{chen2022focalclick} focus on local refinement to improve the mask quality. GPCIS~\cite{zhou2023interactive} formulates IS as a Gaussian process classification to fully propagate click information. 
SimpleClick~\cite{liu2023simpleclick} and iCMFormer~\cite{li2023interactive} achieve superior performance using a Transformer-based architecture that has made brilliant achievements in the field of computer vision~\cite{li2023weakly,li2023acseg,lv2023detrs}.
These methods are all single-granularity output pipelines, ignoring spatial ambiguity.
\smallskip

\noindent \textbf{Multi-granularity Interactive Segmentation.}
A few efforts have been made to tackle the ambiguity in IS. LD~\cite{li2018interactive} proposes to overcome this challenge by using two convolutional networks to select from coarse to precise. 
Recently, the emergence of SAM~\cite{kirillov2023segment} boosts the progress of IS. SAM provides a unified interface to support multiple types of interactions and utilizes the diversity training to attain multi-granularity masks.
Semantic SAM~\cite{li2023semantic} extends the multi-granularity output, but is limited to generating pre-defined segments and only supports a positive click.
These models learn multiple possibilities~\cite{guzman2012multiple} of sparse prompts to dense masks mapping from large-scale multi-granularity annotations~\cite{chen2014detect,he2022partimagenet,kirillov2023segment,lin2014microsoft,shao2019objects365}, which requires expensive data and training costs.
Although the multi-granularity output pipeline alleviates ambiguity, it results in excessive output redundancy and limited scalability.
Unlike previous works, our GraCo resolves ambiguity without redundancy and allows flexible control of prediction granularity without additional manual annotation and extensive training.
\smallskip

\noindent \textbf{Instance and Part Segmentation.}
Instance segmentation is a fundamental task in computer vision that aims to accurately detect and segment each instance.
Instance segmentation has achieved remarkable results after decades of development, and representative works include~\cite{he2017mask,cheng2021per,li2023mask}.
Part segmentation is a sub-task of image segmentation that aims to segment instances into more fine-grained parts.
By identifying the internal structure of objects, part segmentation provides a more comprehensive visual understanding, with typical works including~\cite{li2022panoptic,de2021part}.
Although instance and part segmentation are oriented towards different granularities, both only support segmentation at a fixed granularity and cannot perform human-machine interaction.
Our GraCo supports not only the segmentation of specific parts, but also the flexible manipulation of the granularity level.

\section{The Proposed GraCo}
\label{sec:method}
\begin{figure*}[t]
    \centering
    \includegraphics[width=\textwidth]{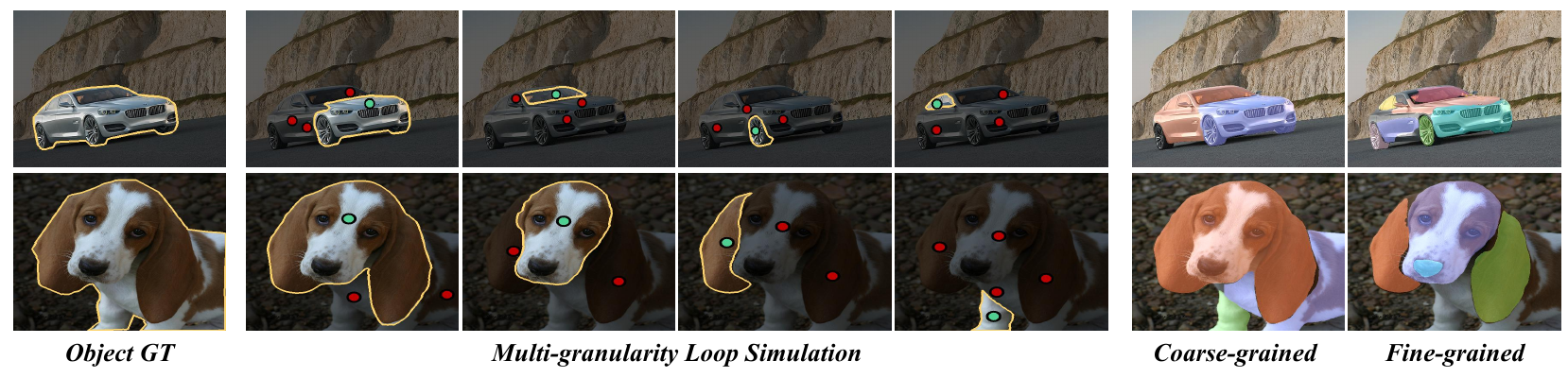}
    \vspace*{-6mm}
    \caption{
        \textbf{Illustration of the multi-granularity loop simulation and visualization of the mask proposals generated by AGG.}
    }
    \vspace*{-4mm}
    \label{fig:proposals}
\end{figure*}
\subsection{Overall Approach}
In this section, we elaborate how to construct the proposed GraCo.
The process of implementing GraCo consists of two stages, \textit{cf.}~\Cref{fig:overview}.
In the first stage, we design an Any-Granularity mask Generator~(AGG), which includes the mask engine and the granularity estimator~(\textit{cf.}~\Cref{subsec:agg}).
The mask engine employs the multi-granularity loop simulation to automatically generate abundant part proposals, and the granularity estimator is responsible for quantifying the granularity of each proposal.
In the second stage, the mask-granularity pairs generated by the previous stage are utilized to perform Granularity-Controllable Learning~(GCL) on the object-level pre-trained IS model~(\textit{cf.}~\Cref{subsec:gcl}).
The details are described as follows.
 
\subsection{Any-Granularity Mask Generator}
\label{subsec:agg}

\noindent \textbf{Mask Engine.}
The core of AGG is the automatic generation of abundant mask-granularity pairs.
To achieve this goal, we exploit the semantic property of the pre-trained IS model to segment local concepts and object parts by simulating appropriate interaction clicks.
Specifically, we first utilize the instance GT as the mask prompt, and randomly select a positive point within the mask to input into the model, marking the object to be parsed.
To drive the mask engine, we design a multi-granularity loop simulation to generate diverse interaction clicks.
At each loop iteration, the click simulator takes a negative click from the current mask and appends it to the click set~(\textit{cf.}~\Cref{fig:proposals}). The current mask is then updated with the model prediction.
Formulaically, given an image $ \mI \in \sR^{h \times w \times 3} $ and a click set $ \mathbfcal{C} $, the positive and negative clicks in set $ \mathbfcal{C} $ are transformed into the disk map $ \mD \in \sR^{h \times w \times 2} $. The object GT is denoted as $ \mG \in \{0, 1\}^{h \times w} $, the IS model $ \mathcal{F}(\cdot) $ outputs the probability for each pixel being foreground. The mask generation process is as follows:
\begin{equation}
\label{equ:initial}
\mY_0 = \mathcal{F}(\mathrm{Fusion}(\mI, \mD_0, \mG)),\ \mY_0 \in [0,1]^{h \times w}, 
\end{equation}
\begin{equation}
\label{equ:recur}
\mY_t = \mathcal{F}(\mathrm{Fusion}(\mI, \mD_t, \mY_{t-1})),\ t=1,2,\dots,N,
\end{equation}
where $\mY_t$ represents the output mask in the $t$-th simulation, $N$ is the number of iterations, and $ \mathrm{Fusion}(\cdot) $ is a fusion operation~(\eg, addition) of all types of features.
In each iteration, we check that the new click is not too close to existing clicks in $ \mathbfcal{C} $, to avoid confusion.
After the loop simulation, the mask engine generates abundant part proposals with diverse granularity.
Furthermore, considering that an entire object consists of multiple parts, we regard the complement within the object of each proposal also as effective parts to increase the diversity of proposals and improve the efficiency of the mask engine.
All proposals are saved after post-processing, which involves morphological processing to eliminate mask holes and connected component filtering to select the connected part.
\smallskip

\noindent \textbf{Granularity Quantification.}
The granularity refers to the level of detail in the segmentation of objects. 
Fine-grained masks furnish rich internal details and part boundaries, while coarse-grained masks provide more general object representations.
To endow the IS model with rational granularity controllability, it is necessary to quantify the granularity consistent with human cognition for each proposal.
Specifically, we consider the granularity quantification from both semantic and scale perspectives.
Semantic granularity is estimated based on the image content covered by the mask, while scale granularity is based on the ratio of the mask in the area to the entire object.
The rationality can be explained as follows. The head of the cat is larger in scale than the crane, as the head accounts for a larger proportion of the cat, but semantically, the two have similar granularity. On the contrary, the feline body can be divided into different granularities, such as individual limbs or specific left and right limbs. Although these two manners possess semantic equivalence, they differ in scale granularity.
\begin{table*}[t]
  \footnotesize
  \centering
  \renewcommand{\arraystretch}{1.15}
  \setlength{\tabcolsep}{2.40pt}
  \begin{tabular*}{\textwidth}{l l | c c | c c | c c | c c | c | c }
    \toprule

    \multirow{2}{*}{\textbf{Method}} & \multirow{2}{*}{\textbf{Backbone}} & 
    \multicolumn{2}{c|}{\textbf{GrabCut}} & \multicolumn{2}{c|}{\textbf{Berkeley}} & \multicolumn{2}{c|}{\textbf{SBD}} & \multicolumn{2}{c|}{\textbf{DAVIS}}
    & \textbf{PascalPart} & \textbf{PartImageNet} \\
    
    \cline{3-12}
    & & \textbf{NoC@85} & \textbf{NoC@90} & \textbf{NoC@85} & \textbf{NoC@90} & \textbf{NoC@85} & \textbf{NoC@90} & \textbf{NoC@85} & \textbf{NoC@90} & \textbf{NoC@85} & \textbf{NoC@85}\\

    \midrule
    \hline

    \rowcolor[gray]{0.9}
    \multicolumn{12}{l}{\emph{Single-granularity Interactive Segmentation}} \\

    DIOS~\cite{xu2016deep} & FCN
    & {5.08} & {6.08} & - & - & {9.22} & {12.80} & {9.03} & {12.58} & - & - \\
    LD~\cite{li2018interactive} & VGG-19
    & {3.20} & {4.79} & - & - & {7.41} & {10.78} & 5.05 & {9.57} & - & - \\
    BRS~\cite{jang2019interactive} & DenseNet
    & {2.60} & {3.60} & - & {5.08} & 6.59 & {9.78} & 5.58 & 8.24 & - & - \\
    FocalClick~\cite{chen2022focalclick} & MiT-B0 
    & 1.66 & 1.90 & - & 3.14 & 4.34 & 6.51 & 5.02 & 7.06 & - & - \\
    FocusCut~\cite{lin2022focuscut} & ResNet-101
    & 1.46 & 1.64 & 1.81 & 3.01 & 3.40 & 5.31 & 4.85 & 6.22 & - & - \\
    PseudoClick~\cite{liu2022pseudoclick} & HRNet-32
    & - & 1.84 & - & 2.98 & - & 5.61 & 4.74 & 6.16 & - & - \\
    CDNet~\cite{chen2021conditional} & ResNet-34
    & 1.86 & 2.18 & 1.95 & 3.27 & 5.18 & 7.89 & 5.00 & 6.89 & 14.95 & 11.96 \\ 
    RITM~\cite{sofiiuk2022reviving} & HRNet-18
    & 1.76 & 2.04 & 1.87 & 3.22 & 3.39 & 5.43 & 4.94 & 6.71 & 10.95 & 9.02 \\
    GPCIS~\cite{zhou2023interactive} & ResNet-50 
    & 1.64 & 1.82 & 1.60 & 2.60 & 3.80 & 5.71 & 4.37 & 5.89 & 10.91 & 8.24 \\
    SimpleClick~\cite{liu2023simpleclick} & ViT-B 
    & 1.40 & {1.54} & {1.44} & {2.46} & {3.28} & {5.24} & {4.10} & {5.48} & 10.97 & 8.58 \\
    SimpleClick~\cite{liu2023simpleclick} & ViT-L 
    & 1.38 & 1.46 & 1.40 & 2.33 & 2.69 & 4.46 & 4.12 & 5.39 & 10.23 & 8.14 \\
    SimpleClick$^\P$~\cite{liu2023simpleclick} & ViT-B 
    & 3.56 & 4.04 & 4.05 & 5.28 & 5.46 & 7.83 & 7.09 & 8.94 & 7.98 & 6.45 \\
    SimpleClick$^\P$~\cite{liu2023simpleclick} & ViT-L
    & 3.86 & 4.48 & 4.43 & 5.66 & 5.59 & 7.92 & 7.62 & 9.46 & 7.28 & 5.81 \\
    
    \rowcolor[gray]{0.9}
    \multicolumn{12}{l}{\emph{Multi-granularity Interactive Segmentation}} \\
    SAM~\cite{kirillov2023segment} & ViT-B
    & 2.42 & 2.72 & 2.21 & 2.96 & 7.22 & 11.05 & 6.13 & 7.88 & 13.89 & 13.32 \\
    SAM~\cite{kirillov2023segment} & ViT-L
    & 1.86 & 1.96 & 1.84 & 2.42 & 5.99 & 9.52 & 4.94 & 6.48 & 13.15 & 11.94 \\
    SAM$^\star$ ~\cite{kirillov2023segment} & ViT-B
    & 1.56 & 1.68 & 1.35 & 1.91 & 6.53 & 10.38 & 4.81 & 6.44 & 13.68 & 12.98 \\
    SAM$^\star$ ~\cite{kirillov2023segment} & ViT-L
    & 1.72 & 1.92 & 1.37 & 2.01 & 5.74 & 9.32 & 5.04 & 6.48 & 13.45 & 12.76 \\
    \thickhline
    \hline

    \rowcolor[gray]{0.9}
    \multicolumn{12}{l}{\emph{Granularity-Controllable Interactive Segmentation (ours)}} \\
    GraCo w/ GT & ViT-B 
    & 1.46 & 1.64 & 1.73 & 2.85 & 3.82 & 5.35 & 5.34 & 7.16 & \underline{6.12} & 6.05 \\
    \rowcolor{aliceblue!80}
    GraCo w/ AGG & ViT-B 
    & \underline{1.34} & \underline{1.46} & \underline{1.37} & \underline{2.21} & \underline{3.44} & \underline{4.89} & \underline{4.44} & \underline{5.72} & {6.38} & \underline{6.01} \\
    \rowcolor{aliceblue!80}
    GraCo w/ GT+AGG & ViT-B 
    & \textbf{1.24} & \textbf{1.36} & \textbf{1.33} & \textbf{2.07} & \textbf{3.22} & \textbf{4.65} & \textbf{4.36} & \textbf{5.49} & \textbf{6.08} & \textbf{5.32} \\
    \hline
    GraCo w/ GT & ViT-L 
    & 1.74 & 1.88 & 1.71 & 2.70 & 3.49 & 4.90 & 5.65 & 7.13 & \textbf{5.81} & 5.34 \\
    \rowcolor{aliceblue!80}
    GraCo w/ AGG & ViT-L 
    & \underline{1.18} & \underline{1.24} & \underline{1.23} & \underline{1.73} & \underline{2.73} & \underline{3.96} & \underline{4.24} & \underline{5.19} & 6.12 & \underline{5.26} \\
    \rowcolor{aliceblue!80}
    GraCo w/ GT+AGG & ViT-L 
    & \textbf{1.18} & \textbf{1.20} & \textbf{1.17} & \textbf{1.61} & \textbf{2.69} & \textbf{3.96} & \textbf{3.87} & \textbf{4.83} & \underline{6.00} & \textbf{4.92} \\

    \bottomrule
  \end{tabular*}
  \vspace*{-2mm}
  \caption{
        \textbf{Comparison with previous methods on both object and part level benchmarks.}
        Single-granularity IS models listed and our GraCo are trained on SBD~\cite{hariharan2011semantic} dataset, and SAM is trained on SA-1B~\cite{kirillov2023segment}.
        All models listed are from official source and use specific data pre-processing pipeline.
        $ \P $ represents fine-tuning the model utilizing the part annotation.
        $ \star $ represents selecting the best matching result from multiple predictions.
        For GraCo, we select the optimal granularity for each instance from 0 to 1 with a step of 0.1 to report the average NoC.
        \textbf{Bold} indicates the best performance and \underline{underlined} the second best.
    }
    \vspace*{-5mm}
  \label{tab:main_sbd}
\end{table*}

\noindent \textbf{Granularity Estimator.}
The granularity estimator is responsible for quantifying the granularity of each proposal $\mP_j^i \in \{0, 1\}^{h \times w} $, where $\mP_j^i$ represents the $i$-th part of object $j$.
We calculate the scale and semantic granularity for each proposal respectively.
The former is directly calculated by dividing the area of the part proposal $\mP_j^i$ by the corresponding instance mask $\mG_j$, and the latter is calculated based on the probability map predicted by the pre-trained IS model.
Specifically, IS model predicts the probability that each pixel belongs to the foreground, and then uses a preset threshold to obtain the binarized mask.
As the threshold increases, the mask shrinks to parts of different scales.
Therefore, we calculate the semantic granularity by the ratio of peak difference $ (\max (\mM_p) - \min (\mM_p)) / (\max (\mM_g) - \min (\mM_g)) $,
where $ \mM $ is the probability map obtained from the pre-trained IS model with a positive click at the center of the mask, $ \mM_p $ and $ \mM_g $ is the probabilities within the proposal $\mP_j^i$ and the corresponding instance mask $\mG_j$.
Formally, the probability map is calculated by~\cref{equ:gra_output}, and the calculation rules for scale and semantic granularity are shown in~\cref{equ:scale} and ~\cref{equ:semantic}.
\begin{equation}
\label{equ:gra_output}
\mM_j^i = \mathcal{F}(\mathrm{Fusion}(\mI, \mD_j^i, \mG_j)),\ \mM_j^i \in \sR^{h \times w},
\end{equation}
\begin{equation}
\label{equ:scale}
\mathcal{G}_{scale}^{i,j} = Area(\mP_j^i) \ / \ Area(\mG_j),
\end{equation}
\begin{equation}
\label{equ:semantic}
\mathcal{G}_{semantic}^{i,j} = \psi(\mM_j^i, \mP_j^i) \ / \ \psi(\mM_j^i, \mG_j), 
\end{equation}
where $Area(\cdot)$ represents the mask area, $\psi(\cdot, \cdot)$ represents the peak difference.
Finally, the granularity of the proposal $\mP_j^i$ is calculated as a linear combination as:
\begin{equation}
\label{equ:gra_sum}
\mathcal{G}^{i,j} = (1 - \lambda) \cdot \mathcal{G}_{scale}^{i,j} + \lambda \cdot \mathcal{G}_{semantic}^{i,j},
\end{equation}
where $\lambda$ represents the weight coefficient, which is set to 0.5 in the experiments.

\subsection{Granularity-Controllable Learning}
\label{subsec:gcl}

\noindent \textbf{Granularity Embedding.}
We transform the granularity into the learnable embedding as an additional prompt to the IS model.
According to~\Cref{equ:scale} and ~\Cref{equ:semantic}, it is apparent that the granularity fall within the range of [0,1]. 
Therefore, we discretize the interval from 0 to 1 into $B$ bins and establish a table that maps the discrete granularities to high-dimensional embeddings.
The prompts, including granularity, clicks and mask, are integrated with the image embedding and jointly fed into the feature extractor.
\smallskip

\noindent \textbf{Proposal Sampling and Training.}
Considering the uneven granularity distribution of mask-granularity pairs generated by AGG, we formulate the sampling probability of each mask as an inversely proportional function of the ratio of the corresponding granularity in the proposal database to improve the training stability.
For training, the IS model utilizes the iterative sampling strategy~\cite{sofiiuk2022reviving,liu2023simpleclick}. The segmentation of the previous iteration step serves as the mask prompt for the model and we feed an empty mask for the first iteration.
The iterative sampling strategy achieves a high-level of consistency in simulating the user behaviour, thereby improving performance.
We take the Normalized Focal Loss~(NFL) following~\cite{liu2023simpleclick,li2023interactive} for training.
\smallskip

\noindent \textbf{LoRA Technology.}
We utilize LoRA technology~\cite{hu2021lora} to facilitate the object-level pre-trained IS model in efficiently comprehending granularity controllability while preserving its primary performance.
For the feature extractor with a weight matrix $\mW \in \mathrm{R}^{d \times d}$, we maintain the $ \mW $ frozen while learning a new weight matrix $ \mB\mA $.
Formulaically, given a feature extractor $\mathcal{E}(\cdot)$ and input $\boldsymbol{x}$, the forward process is represented as:
\begin{equation}
\label{equ:lora}
\mathcal{E}(\boldsymbol{x}) = \mW \boldsymbol{x} + \mB\mA \boldsymbol{x},
\end{equation}
where $\mB \in \mathrm{R}^{d \times r}$ and $\mA \in \mathrm{R}^{r \times d}$. The rank $r$ is typically lower than the dimension $d$ to reduce the computational cost.
For implementation, $\mA$ employs Gaussian initialization while $\mB$ initializes with zero, ensuring that $\mB\mA$ is a zero matrix at the start of fine-tuning.
We apply LoRA to the projection layers of $\mQ$ and $\mK$ in each attention block.

\section{Experiments}
\label{sec:exp}
\subsection{Experimental Settings}
\begin{figure*}[t]
    \centering
    \includegraphics[width=\textwidth]{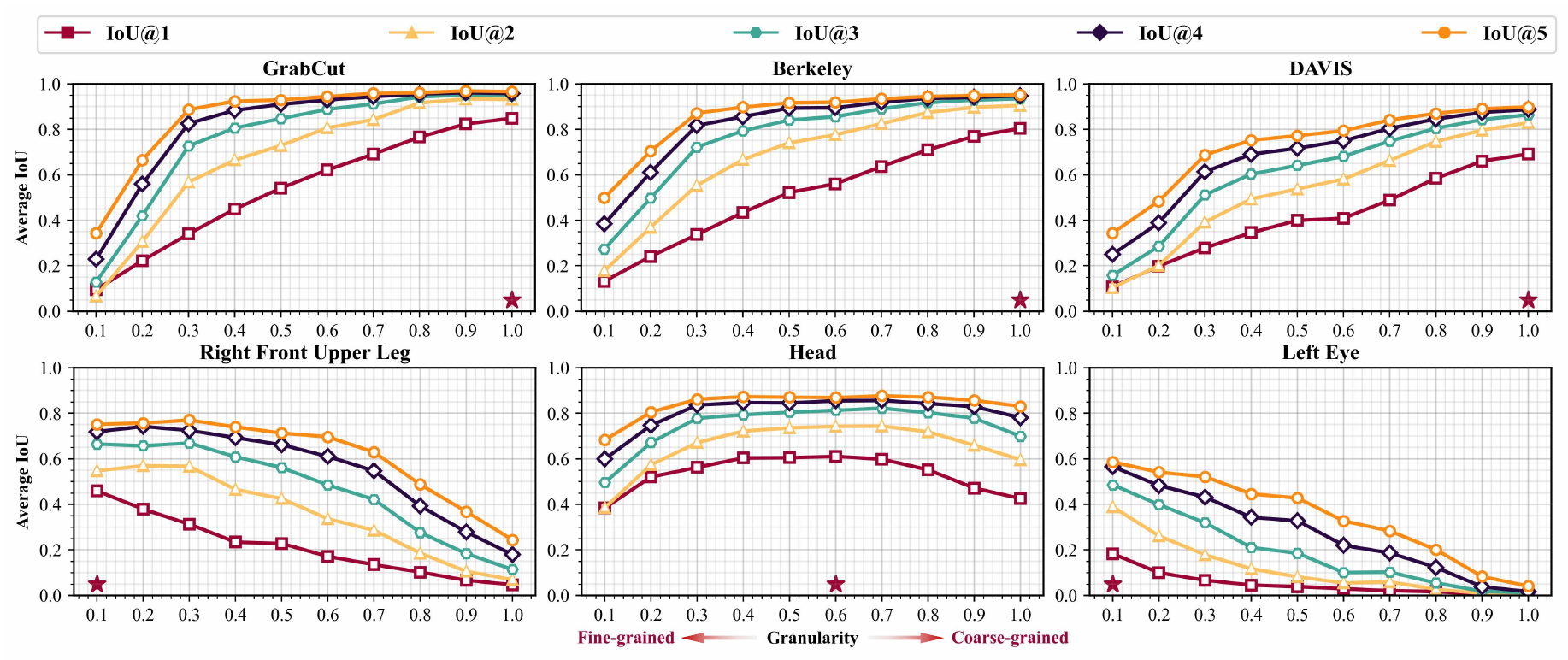}
    \vspace*{-7mm}
    \caption{
    \textbf{Verification of the granularity controllability.}
    We calculate IoU@k under different granularities to plot IoU-granularity curves.
    The optimal granularity (marked by the \textcolor[rgb]{0.631, 0, 0.208}{red} star) of the objects is about 1.0, while for the parts of the cow from PascalPart~\cite{chen2014detect} it is different.
    }
    \vspace*{-5mm}
    \label{fig:ious_object}
\end{figure*}
\noindent \textbf{Dataset.}
To demonstrate the performance of the IS model in multi-granularity scenarios, we utilize object and part level benchmarks for evaluation.
For the object-level, we conduct evaluation on four commonly used benchmarks: \textbf{GrabCut}~\cite{rother2004grabcut}, \textbf{Berkeley}~\cite{mcguinness2010comparative}, \textbf{SBD}~\cite{hariharan2011semantic}, \textbf{DAVIS}~\cite{perazzi2016benchmark}. 
For the part-level, we utilize two part segmentation datasets: \textbf{PascalPart}~\cite{chen2014detect} and \textbf{PartImageNet}~\cite{he2022partimagenet}.
Note that we train our GraCo on SBD and remove samples from the PascalPart validation set that belong to the SBD training set. See the Appendix for a detailed description of these datasets.

\smallskip

\noindent \textbf{Implementation Details.}
We build our GraCo based on SimpleClick~\cite{liu2023simpleclick}, which consists of two patch embedding modules for image and click map respectively~(we introduce an extra granularity embedding for our GraCo), a ViT~\cite{dosovitskiy2020image} backbone initialized with MAE~\cite{he2022masked}, a simple feature pyramid~\cite{li2022exploring}, and an MLP segmentation head.
The IS model employed in AGG is SimpleClick with ViT-Base.
The multi-granularity loop simulation iterations for each instance are randomly selected from a range of 3 to 6.
For LoRA~\cite{hu2021lora}, the rank is set to 8 and the discretization interval for granularity is set to 0.1.
We set the maximum number of iterative clicks to 3 follow~\cite{liu2023simpleclick}.
We train the GraCo for 55 epochs using the Adam~\cite{kingma2014adam} optimizer with a learning rate of 5e-5, which decays by a factor of 10 at 50 epochs.
For inference, we set the threshold for binarizing the prediction to 0.5 and use the same data augmentation as~\cite{li2023multi}.

\smallskip

\begin{figure*}[t]
    \centering
    \includegraphics[width=\textwidth]{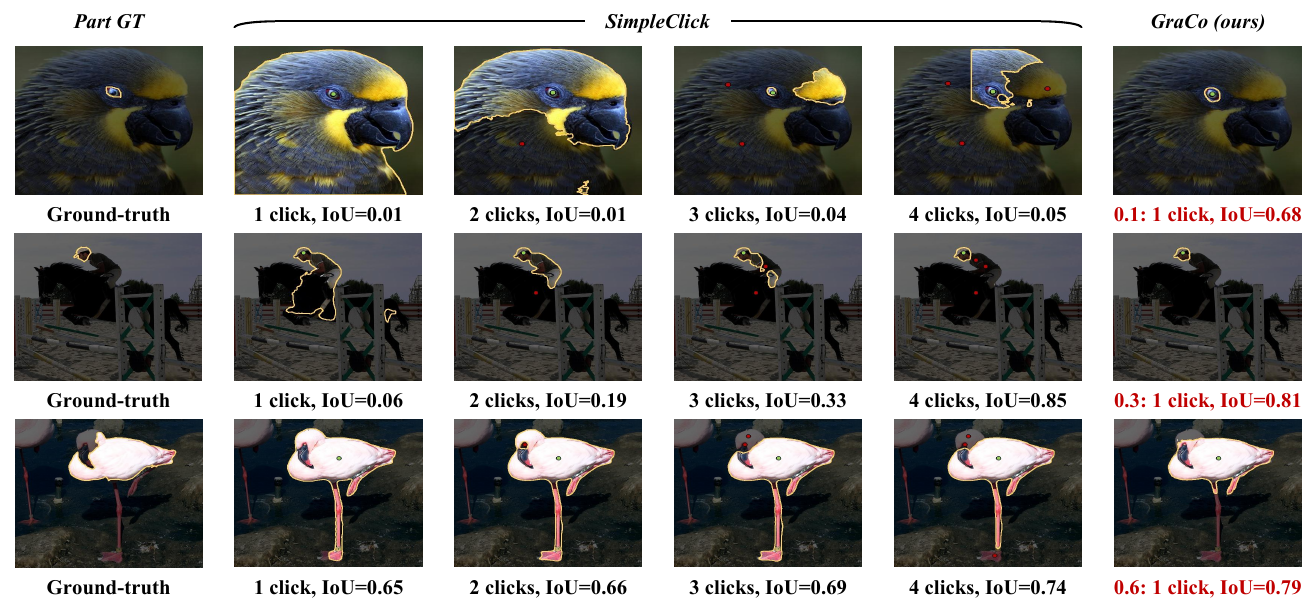}
    \vspace*{-7mm}
    \caption{
        \textbf{Visualization of interactive segmentation on part GT using SimpleClick}~\cite{liu2023simpleclick} \textbf{and our GraCo.}
        We note the input granularity for our GraCo, which is roughly estimated based on human cognition.
    }
    \vspace*{-4mm}
    \label{fig:vis}
\end{figure*}
\noindent \textbf{Evaluation Protocol.}
We conduct the evaluation following the standard protocol of previous click-based IS methods~\cite{xu2016deep,chen2021conditional,sofiiuk2022reviving,chen2022focalclick,lin2022focuscut,liu2023simpleclick}.
Specifically, the first positive click is sampled in the center of the object, while the subsequent clicks are derived from the largest error region by comparing the current mask with the GT.
For the metrics, we adopt the Number of Click (NoC) to evaluate the performance, which counts the average number of clicks required to achieve a fixed Intersection over Union (IoU), with lower values indicating better performance.
We set two commonly used target IoU thresholds (85\% and 90\%, denoted as NoC@85 and NoC@90 respectively) and 20 clicks as the upper bound for interaction, which are same with previous works~\cite{sofiiuk2022reviving,liu2023simpleclick,li2023interactive}.
Moreover, the IoU-granularity curves are drawn to verify the granularity controllability of our GraCo.
We also calculate the average IoU of the first click, and the results are shown in the Appendix 2.1.

\subsection{Main Results and Analysis}
\noindent \textbf{Comparison with Previous Method.}
We compare our results with previous single and multiple granularity IS methods on four object-level benchmarks and two part-level benchmarks. 
Note that we report NoC@85 and NoC@90 for the object-level benchmarks and only NoC@85 for the part-level benchmarks. The reason is that multi-granularity parts are more difficult to segment than objects.
As a result, it is challenging to achieve an IoU of up to 90\% within 20 clicks.
The experimental results are shown in~\Cref{tab:main_sbd}.
We present the results of single-granularity models equipped with different backbones trained on SBD~\cite{hariharan2011semantic}, alongside the results of the multi-granularity model~(\ie, SAM) trained on SA-1B~\cite{kirillov2023segment}. We utilize the official models and retain their specific data pre-processing pipeline for evaluation.
For our GraCo, we present the performance using the mask proposals generated by our AGG~(denoted as GraCo w/ AGG in~\Cref{tab:main_sbd}).
Based on the results, single-granularity IS methods show satisfactory performance in object-level benchmarks, but poor performance in handling the part-level, and the multi-granularity method perform poorly at both levels.
In contrast, our GraCo w/ AGG achieves superior performance on all benchmarks at both levels.

In addition, we fine-tune SimpleClick~\cite{liu2023simpleclick} and our GraCo utilizing the training set of SBD~\cite{hariharan2011semantic}  with part annotations from PascalPart~(denoted as SimpleClick$^\P$ and GraCo w/ GT).
The results of SimpleClick$^\P$ indicate that fine-tuning the model with part annotations not only weakens the object-level segmentation performance, but also achieves a marginal improvement at the part-level.
However, our GraCo w/ GT using the proposed GCL strategy achieves significant performance improvements over vanilla SimpleClick, demonstrating the effectiveness of GCL.

\noindent \textbf{Failure Analysis of SAM.} 
SAM~\cite{kirillov2023segment}, a representative of multi-granularity IS methods, does not achieve ideal results in~\Cref{tab:main_sbd}, which is below our expectations.
Upon our analysis, we find that SAM has a bias towards segmenting small components on object-level benchmarks even when producing multiple masks.
This factor causes SAM to require more clicks to reach the IoU thresholds, resulting in unsatisfactory NoC.
Furthermore, the mask distribution of the selected part-level benchmarks deviates from its training set, exposing its limited generalization.
To substantiate this claim, we evaluate the performance of SAM, SimpleClick~\cite{liu2023simpleclick}, and our GraCo using the first 1000 images from the SA-1B~\cite{kirillov2023segment} as a dedicated test subset in~\Cref{tab:sa1b}. 
Considering that each image in SA-1B contains an average of 100 masks, covering diverse granularities and overlapping, we select five non-overlapping masks for each image (selecting 4987 masks in total) for evaluation.
We conclude that SimpleClick performs poorly on such a multi-granularity benchmark, while SAM achieves excellent performance because it is a subset of its training set, which is in line with our expectations.
Our GraCo achieves comparable NoC@90 metrics to SAM, while significantly outperforming SimpleClick. This demonstrates the robust generalization and excellent performance of GraCo in multi-granularity segmentation.
Furthermore, we also calculate the IoU@1 on all benchmarks. 
We find that SAM achieves superior performance when producing multiple masks, providing an excellent user experience.
The detailed results are shown in the Appendix 2.1.

\begin{table}[!t]
\centering
\footnotesize
\renewcommand{\arraystretch}{1.15}
\setlength{\tabcolsep}{4.8pt}
\begin{tabular*}{\linewidth}{ ll| c | c | c }
  \toprule
   \multirow{2}{*}{\textbf{Method}} & \multirow{2}{*}{\textbf{Backbone}} & 
    \multicolumn{3}{c}{\textbf{SA-1B}~\cite{kirillov2023segment}} \\ 
    \cline{3-5}
    &  & \textbf{NoC@85$ \downarrow $} & \textbf{NoC@90$ \downarrow $} & \textbf{IoU@1$ \uparrow $}  \\ 
  \midrule
  \midrule
  SimpleClick~\cite{liu2023simpleclick} & ViT-B & 5.56 & 7.29 & 0.22 \\
  SAM~\cite{kirillov2023segment} & ViT-B & \underline{2.93} & 5.19 & \underline{0.78} \\
  SAM$^\star$~\cite{kirillov2023segment} & ViT-B & \textbf{2.46} & \underline{4.42} & \textbf{0.88} \\
  \rowcolor{aliceblue}
  GraCo w/ AGG & ViT-B & 3.39 & \textbf{4.29} & 0.61 \\
  \hline
  SimpleClick~\cite{liu2023simpleclick} & ViT-L & 4.98 & 6.74 & 0.29 \\
  SAM~\cite{kirillov2023segment} & ViT-L & \underline{1.99} & \underline{3.31} & \underline{0.81} \\
  SAM$^\star$~\cite{kirillov2023segment} & ViT-L & \textbf{1.77} & \textbf{2.97} & \textbf{0.91} \\
  \rowcolor{aliceblue}
  GraCo w/ AGG & ViT-L & 3.10 & 3.96 & 0.65 \\
  \bottomrule
\end{tabular*}
\vspace*{-2mm}
\caption{\textbf{Experimental results on the first 1000 images of SA-1B~\cite{kirillov2023segment}.} $ \star $, \textbf{Bold} and \underline{underlined} are the same as~\Cref{tab:main_sbd}.}
\vspace*{-6mm}
\label{tab:sa1b}
\end{table}
\begin{table*}[!t]
\footnotesize
  \centering
  \renewcommand{\arraystretch}{1.15}
  \setlength{\tabcolsep}{2.78pt}
\begin{tabular*}{\textwidth}{c c | c c c | c c c | c c c | c c }
  \toprule
   \multirow{2}{*}{\textbf{LoRA}} & \textbf{Granularity} & \multicolumn{3}{c|}{\textbf{GrabCut}} & \multicolumn{3}{c|}{\textbf{Berkeley}} & \multicolumn{3}{c|}{\textbf{SBD}} & \multicolumn{2}{c}{\textbf{PascalPart}} \\ 
   \cline{3-13}
 & \textbf{Embedding} & \textbf{NoC@85$ \downarrow $} & \textbf{NoC@90$ \downarrow $} & \textbf{IoU@1$ \uparrow $} & \textbf{NoC@85$ \downarrow $} & \textbf{NoC@90$ \downarrow $} & \textbf{IoU@1$ \uparrow $} & \textbf{NoC@85$ \downarrow $} & \textbf{NoC@90$ \downarrow $} & \textbf{IoU@1$ \uparrow $} & \textbf{NoC@85$ \downarrow $} & \textbf{IoU@1$ \uparrow $}\\
  \midrule
  \midrule
  - & - & 3.56 & 4.04 & 0.47 & 4.05 & 5.28 & 0.43 & 5.46 & 7.83 & 0.42 & 7.98 & 0.48 \\
  \checkmark & - & 3.24 & 3.68 & 0.45 & 3.67 & 4.97 & 0.41 & 4.66 & 6.68 & 0.48 & 8.68 & 0.43 \\
  - & \checkmark & 2.14 & 2.52 & 0.79 & 1.90 & \textbf{2.78} & 0.79 & 4.20 & 5.99 & 0.64 & \textbf{5.84} & \textbf{0.59} \\
  \rowcolor{aliceblue}
  \checkmark & \checkmark & \textbf{1.46} & \textbf{1.64} & \textbf{0.86} & \textbf{1.73} & 2.85 & \textbf{0.80} & \textbf{3.82} & \textbf{5.35} & \textbf{0.66} & 6.12 & 0.52 \\
  \bottomrule
\end{tabular*}
\vspace*{-1mm}
\caption{\textbf{Results of ablation study on GCL.} We utilize SimpleClick~\cite{liu2023simpleclick} with ViT-B to train on SBD~\cite{hariharan2011semantic} with part annotations.}
\vspace*{-3mm}
\label{tab:gis_finetuning}
\end{table*}
\smallskip

\noindent \textbf{Gains from AGG.}
We utilize part annotations, mask proposals generated by AGG, and the combination of both to perform the GCL strategy, corresponding to GraCo w/ GT, GraCo w/ AGG, and GraCo w/ GT+AGG in~\Cref{tab:main_sbd}.
Taking advantage of the any-granularity part proposals generated by AGG, GraCo w/ AGG performs better than GraCo w/ GT on all benchmarks except PascalPart~\cite{chen2014detect}.
We argue that this is due to the limited number of manual annotations and the existence of granularity variance, which cannot cover arbitrary granularities, resulting in sub-optimal generalization.
In contrast, AGG automatically generates abundant any-granularity masks, thereby facilitating the IS model in capturing granularity controllability.
Moreover, the results of GraCo w/ GT+AGG are superior to both GraCo w/ GT and GraCo w/ AGG, further demonstrating that the proposals generated by AGG offer a greater level of granularity abundance than GT and serve as an effective supplement.

\smallskip

\noindent \textbf{Granularity Controllability Analysis.} 
To verify the granularity controllability of our GraCo, we calculate the IoU@k at different granularities and plot the IoU-granularity curves~(\textit{cf.}~\Cref{fig:ious_object}).
Based on the granularity definition, 1.0 represents object-level segmentation, and the closer to 0, the finer the prediction granularity.
For three object-level benchmarks, IoU@k increases with increasing granularity, especially IoU@1, which is as expected.
For the part-level scenario, we randomly select three part categories belonging to the cow category for validation.
For highly detailed parts such as the right front upper leg and left eye, GraCo performs optimally at a granularity of 0.1. For coarse-grained parts such as the head, the optimal granularity for GraCo is around 0.6.
The part-level results further demonstrate that our GraCo possess granularity controllability consistent with human cognition.

\smallskip

\noindent \textbf{Qualitative Results.}
\Cref{fig:vis} shows the qualitative results using SimpleClick~\cite{liu2023simpleclick} and our GraCo on some segmentation examples.
We randomly select several parts from PascalPart~\cite{chen2014detect} annotations and automatically generate the next click according to the evaluation protocol.
We find that SimpleClick requires multiple clicks to segment the desired mask in multi-granularity scenarios.
In contrast, our GraCo requires only a single click to match expectations well based on roughly estimated input granularity.
This demonstrates the flexibility of our GraCo to adapt to diverse scenarios.

\smallskip

\subsection{Ablation Study}
\noindent \textbf{Granularity-Controllable Learning.}
To demonstrate the effectiveness of the GCL strategy, we evaluate the contributions of its two key components, \ie, granularity embedding and low-rank adaptation.
Specifically, we conduct experiments including removing granularity embedding, removing LoRA (\ie, full parameter fine-tuning), and removing both simultaneously~(\textit{cf.}~\Cref{tab:gis_finetuning}).
We conclude that incorporating granularity embedding effectively enhances the performance, whereas the LoRA technology preserves the original performance of the pre-trained model. 
More detailed ablation studies are provided in Appendix 2.2.

\smallskip

\begin{figure}
  \centering
  \begin{subfigure}{0.49\linewidth}
  \includegraphics[width=\linewidth]{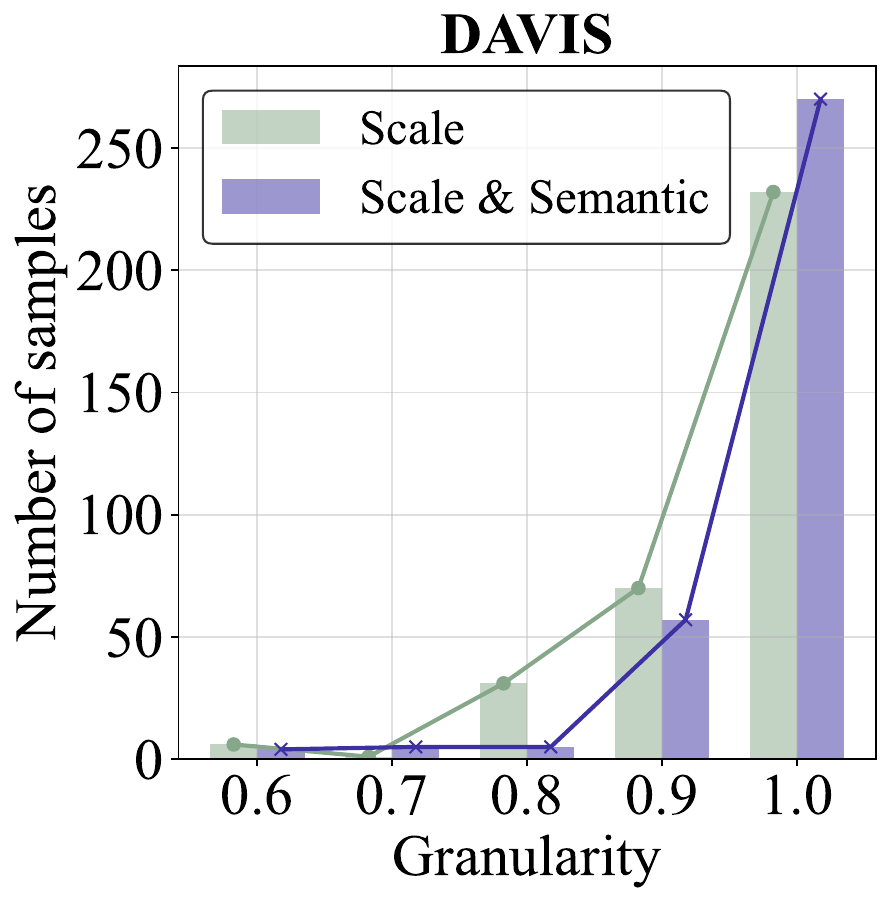} 
    \label{fig:gra_define-davis}
  \end{subfigure}
  \hfill
  \begin{subfigure}{0.495\linewidth}
    \includegraphics[width=\linewidth]{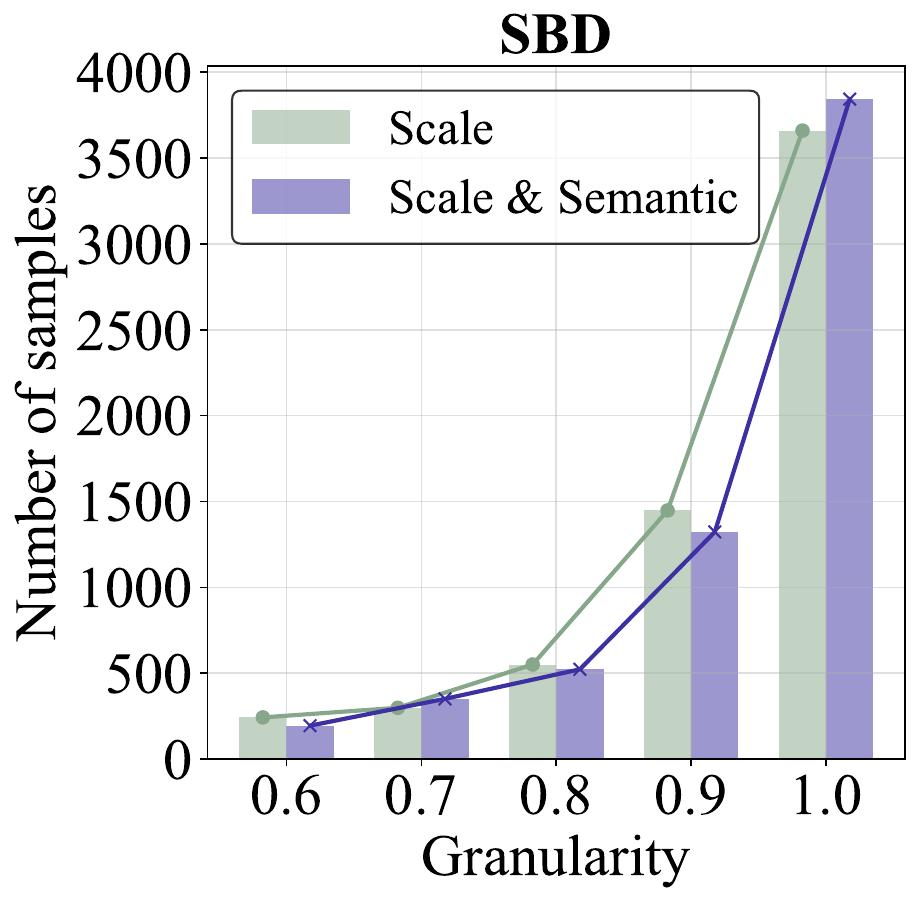} 
    \label{fig:gra_define-sbd}
  \end{subfigure}
  \vspace*{-9mm}
  \caption{\textbf{Frequency distribution of optimal granularity.}}
  \label{fig:gra_define}
  \vspace*{-6mm}
\end{figure}

\noindent \textbf{Granularity Definition.} 
To demonstrate the necessity of both semantic and scale granularity, we conduct experiments in two settings: with scale granularity only, and with both scale and semantic granularity.
We plot a histogram and line graph to display the frequency distribution of optimal granularity on two object-level benchmarks, \ie, DAVIS~\cite{perazzi2016benchmark} and SBD~\cite{hariharan2011semantic}, \Cref{fig:gra_define}.
We conclude that the optimal granularity tends to be skewed to 1.0 when employing both scale and semantic granularity. This aligns with the granularity definition for the whole instance.
Moreover, we quantitatively evaluate the performance of the two settings on part-level benchmarks in Appendix 2.2, which demonstrates the necessity of the two types of granularity.

\section{Conclusion}
\label{sec:conclusion}
In this work, we propose a novel paradigm for interactive segmentation that allows users to control the segmentation granularity to resolve ambiguity.
Our GraCo fine-tunes the pre-trained IS model to endow it with granularity controllability without requiring additional manual annotation, providing a non-redundant, low-cost and highly flexible solution to address spatial ambiguity.
Excellent experimental results demonstrate the effectiveness and generalization of our method, and the granularity controllability analysis confirms the consistency of the model with human cognition.
We hope that our exploration will open up new avenues for resolving ambiguity in pixel-level interactive AI systems.
\smallskip

\noindent \textbf{Acknowledgements.} This work was supported in part by the National Key R\&D Program of China (No. 2022ZD0118201), Natural Science Foundation of China (No. 61972217, 32071459, 62176249, 62006133, 62271465), and the Shenzhen Medical Research Funds in China (No. B2302037).

{
    \small
    \bibliographystyle{ieeenat_fullname}
    \bibliography{main}
}

\clearpage
\setcounter{page}{1}
\maketitleappendix

\renewcommand{\thefootnote}{\fnsymbol{footnote}}

\renewcommand{\thetable}{\Alph{table}}
\renewcommand{\theequation}{\Alph{equation}}
\renewcommand{\thefigure}{\Alph{figure}}

\setcounter{table}{0}
\setcounter{section}{0}
\setcounter{figure}{0}
\setcounter{equation}{0}

\section{Limitations} 
In this work, we introduce \textbf{Gra}nularity-\textbf{Co}ntrollable interactive segmentation~(\textbf{GraCo}) that allows users to control the segmentation granularity to resolve ambiguity.
Although we develop a novel and flexible paradigm and achieve inspiring results, the proposed method still has some limitations:
(\romannumeral1). Due to the randomness in the interaction signals generated by the multi-granularity loop simulation in the any-granularity mask generator, which causes the object-level pre-trained IS model to generate semantically inconsistent parts or noisy boundaries, providing inaccurate granularity-controllability guidance.
(\romannumeral2). Considering the variance in the computational cost of running the mask engine at different granularities, we choose to generate proposals offline to improve the efficiency of parallel computing. As a result, there is a trade-off between storage space and granularity abundance. The online fine-tuning paradigm of granularity-controllability is a future exploration to overcome this limitation.

\section{Additional Experiments and Analysis}

\subsection{IoU@1 Analysis} 
\label{subsec:iou_analysis}

\begin{table*}[!ht]
\centering
\footnotesize
\renewcommand{\arraystretch}{1.15}
\setlength{\tabcolsep}{8.25pt}
\begin{tabular*}{\textwidth}{ ll| l | l | l | l | l | l }
  \toprule
   \textbf{Method} & \textbf{Backbone} & \multicolumn{1}{c|}{\textbf{GrabCut}} & \multicolumn{1}{c|}{\textbf{Berkeley}} & \multicolumn{1}{c|}{\textbf{SBD}} & \multicolumn{1}{c|}{\textbf{DAVIS}} & \multicolumn{1}{c|}{\textbf{PascalPart}} & \multicolumn{1}{c}{\textbf{PartImageNet}} \\ 
  \midrule
  \midrule
  SimpleClick~\cite{liu2023simpleclick} & ViT-B & 0.90 & 0.85 & 0.74 & 0.76 & 0.17 & 0.30 \\
  SimpleClick$^\P$~\cite{liu2023simpleclick} & ViT-B & 0.47 \decrease{0.43} & 0.43 \decrease{0.42} & 0.42 \decrease{0.32} & 0.31 \decrease{0.45} & 0.48 \increase{0.31} & 0.49 \increase{0.19} \\
  SimpleClick~\cite{liu2023simpleclick} & ViT-L & 0.91 & 0.84 & 0.82 & 0.78 & 0.18 & 0.30 \\
  SimpleClick$^\P$~\cite{liu2023simpleclick} & ViT-L & 0.48 \decrease{0.43} & 0.46 \decrease{0.38} & 0.46 \decrease{0.36} & 0.38 \decrease{0.40} & 0.53 \increase{0.35} & 0.54 \increase{0.24} \\
  \hline
  SAM~\cite{kirillov2023segment} & ViT-B & 0.55 & 0.56 & 0.45 & 0.41 & 0.43 & 0.42 \\
  SAM$^\star$~\cite{kirillov2023segment} & ViT-B & 0.90 \increase{0.35} & 0.88 \increase{0.32} & 0.75 \increase{0.30} & 0.74 \increase{0.33} & 0.57 \increase{0.14} & 0.55 \increase{0.13} \\
  SAM~\cite{kirillov2023segment} & ViT-L & 0.61 & 0.61 & 0.50 & 0.45 & 0.44 & 0.42 \\
  SAM$^\star$~\cite{kirillov2023segment} & ViT-L & 0.94 \increase{0.33} & 0.90 \increase{0.29} & 0.80 \increase{0.30} & 0.78 \increase{0.33} & 0.57 \increase{0.13} & 0.56 \increase{0.14} \\
    \hline
  \rowcolor{aliceblue}
  GraCo w/ GT & ViT-B & 0.86 & 0.80 & 0.66 & 0.62 & 0.52 & 0.53 \\
  \rowcolor{aliceblue}
  GraCo w/ AGG & ViT-B & 0.89 \increase{0.03} & 0.84 \increase{0.04} & 0.72 \increase{0.06} & 0.70 \increase{0.08} & 0.53 \increase{0.01} & 0.55 \increase{0.02} \\
  \rowcolor{aliceblue}
  GraCo w/ GT & ViT-L & 0.81 & 0.76 & 0.66 & 0.56 & 0.56 & 0.55 \\
  \rowcolor{aliceblue}
  GraCo w/ AGG & ViT-L & 0.93 \increase{0.12} & 0.89 \increase{0.13} & 0.81 \increase{0.15} & 0.75 \increase{0.19} & 0.55 \decrease{0.01} & 0.58 \increase{0.03} \\
  \bottomrule
\end{tabular*}
\vspace*{-2mm}
\caption{\textbf{IoU@1 Analysis on both object and part level benchmarks.} $ \P $ represents fine-tuning the model utilizing the part annotation, and $ \star $ represents selecting the best matching result from multiple predictions. SimpleClick~\cite{liu2023simpleclick} and our GraCo are trained on SBD~\cite{hariharan2011semantic} and SAM are trained on SA-1B~\cite{kirillov2023segment}. SimpleClick and SAM are from official models and use specific data pre-processing pipeline.}
\vspace*{-1mm}
\label{tab:iou1}
\end{table*}

\begin{table*}[!ht]
\footnotesize
  \centering
  \renewcommand{\arraystretch}{1.15}
  \setlength{\tabcolsep}{3.68pt}
\begin{tabular*}{\textwidth}{c | c c c | c c c | c c c | c c }
  \toprule
   \multirow{2}{*}{\textbf{Sampling}} & \multicolumn{3}{c|}{\textbf{GrabCut}} & \multicolumn{3}{c|}{\textbf{Berkeley}} & \multicolumn{3}{c|}{\textbf{SBD}} & \multicolumn{2}{c}{\textbf{PascalPart}} \\ 
   \cline{2-12}
   & \textbf{NoC@85$ \downarrow $} & \textbf{NoC@90$ \downarrow $} & \textbf{IoU@1$ \uparrow $} & \textbf{NoC@85$ \downarrow $} & \textbf{NoC@90$ \downarrow $} & \textbf{IoU@1$ \uparrow $} & \textbf{NoC@85$ \downarrow $} & \textbf{NoC@90$ \downarrow $} & \textbf{IoU@1$ \uparrow $} & \textbf{NoC@85$ \downarrow $} & \textbf{IoU@1$ \uparrow $}\\
  \midrule
  \midrule
  \textbf{Uniform} & 1.46 & 1.52 & 0.86 & 1.41 & 2.29 & 0.83 & 3.49 & 4.93 & 0.70 & 6.44 & 0.52 \\
  \rowcolor{aliceblue}
  \textbf{Inverse-prop.} & \textbf{1.34} & \textbf{1.46} & \textbf{0.89} & \textbf{1.37} & \textbf{2.21} & \textbf{0.84} & \textbf{3.44} & \textbf{4.89} & \textbf{0.72} & \textbf{6.38} & \textbf{0.53} \\
  \bottomrule
\end{tabular*}
\vspace*{-2mm}
\caption{\textbf{Results of ablation study on proposal sampling.}}
\vspace*{-1mm}
\label{tab:sampling}
\end{table*}

\begin{table*}[!ht]
\centering
\footnotesize
\renewcommand{\arraystretch}{1.15}
\setlength{\tabcolsep}{6.35pt}
\begin{tabular*}{\textwidth}{c | c c | c c | c c | c c | c | c }
  \toprule
   {\textbf{LoRA}} & 
    \multicolumn{2}{c|}{\textbf{GrabCut}} & \multicolumn{2}{c|}{\textbf{Berkeley}} & \multicolumn{2}{c|}{\textbf{SBD}} & \multicolumn{2}{c|}{\textbf{DAVIS}}
    & \textbf{PascalPart} & \textbf{PartImageNet} \\
    
    \cline{2-11}
    {\textbf{Rank}} & \textbf{NoC@85} & \textbf{NoC@90} & \textbf{NoC@85} & \textbf{NoC@90} & \textbf{NoC@85} & \textbf{NoC@90} & \textbf{NoC@85} & \textbf{NoC@90} & \textbf{NoC@85} & \textbf{NoC@85}\\
  \midrule
  \midrule

  \textbf{4} & 1.36 & 1.48 & 1.43 & 2.25 & 3.48 & 4.93 & 4.62 & 5.84 & 6.47 & 6.03 \\
   \rowcolor{aliceblue}
  \textbf{8} & \underline{1.34} & \underline{1.46} & \textbf{1.37} & \textbf{2.21} & \underline{3.44} & \underline{4.89} & \underline{4.44} & \underline{5.72} & \textbf{6.38} & \textbf{6.01} \\

  \textbf{16} & \textbf{1.32} & \textbf{1.44} & 1.40 & \underline{2.23} & 3.45 & 4.90 & 4.68 & 5.85 & \underline{6.39} & \underline{6.03} \\

  \textbf{32} & 1.32 & 1.44 & 1.37 & 2.24 & \textbf{3.40} & \textbf{4.85} & \textbf{4.41} & \textbf{5.70} & 6.42 & 6.03 \\
  \bottomrule
\end{tabular*}
\vspace*{-2mm}
\caption{\textbf{Ablation study on LoRA.} We train our GraCo on the same AGG-generated proposals with different ranks of the LoRA. We utilize ViT-B as the backbone. \textbf{Bold} indicates the best performance and \underline{underlined} the second best.}
\vspace*{-2mm}
\label{tab:lora}
\end{table*}
\begin{table}[!ht]
\footnotesize
  \centering
  \renewcommand{\arraystretch}{1.15}
  \setlength{\tabcolsep}{4.87pt}
\begin{tabular*}{\linewidth}{c | c c | c c }
  \toprule
   \textbf{Granularity} & \multicolumn{2}{c|}{\textbf{PascalPart}} & \multicolumn{2}{c}{\textbf{PartImageNet}} \\ 
   \cline{2-5}
 \textbf{Definition} & \textbf{NoC@85$ \downarrow $} & \textbf{IoU@1$ \uparrow $} & \textbf{NoC@85$ \downarrow $} & \textbf{IoU@1$ \uparrow $} \\
  \midrule
  \midrule
  \textbf{Scale-only} &  6.43 & 0.52 & 6.08 & 0.54 \\
  \rowcolor{aliceblue}
  \textbf{Scale \& Semantic} & \textbf{6.38} &\textbf{ 0.53} & \textbf{6.01} & \textbf{0.55} \\
  \bottomrule
\end{tabular*}
\vspace*{-1mm}
\caption{\textbf{Results of ablation study on granularity definition.}}
\vspace*{-5mm}
\label{tab:granularity}
\end{table}

Considering that the segmentation mask after the first click directly affects the user experience, we evaluate the IoU@1 of the IS methods.
As shown in~\Cref{tab:iou1}, we evaluate the IoU@1 of SimpleClick~\cite{liu2023simpleclick}, SAM~\cite{kirillov2023segment} and our GraCo.
For SimpleClick, we report the results of the pre-trained model and the model fine-tuned with part annotations.
From the results, we conclude that fine-tuning with part annotations leads to a significant decrease in IoU@1 on object-level benchmarks. In contrast, the results on part-level benchmarks are effectively improved, indicating that the model tends to perform fine-grained part segmentation after fine-tuning.
For SAM, we present the results for single-output and multi-output~(default $3$) respectively.
We observe that SAM exhibits excellent performance. Specifically, the first click performance of SAM is significantly superior than SimpleClick, especially when selecting the optimal mask from multiple outputs for each instance.
Moreover, the IoU@1 obtained by multi-output outperforms single-output considerably, as denoted by the green-highlighted increment.
This enhances SAM's user experience.
For our GraCo, we present the results of fine-tuning with part annotations and AGG-generated mask proposals respectively. 
We observe that GraCo w/ AGG is superior than GraCo w/ GT. We argue that this is because AGG generates a wealth of mask proposals to cover a wider range of granularity. 
Our GraCo achieves comparable first click performance to SAM on all benchmarks at a low cost.

\begin{figure*}[h]
    \centering
    \includegraphics[width=0.96\textwidth]{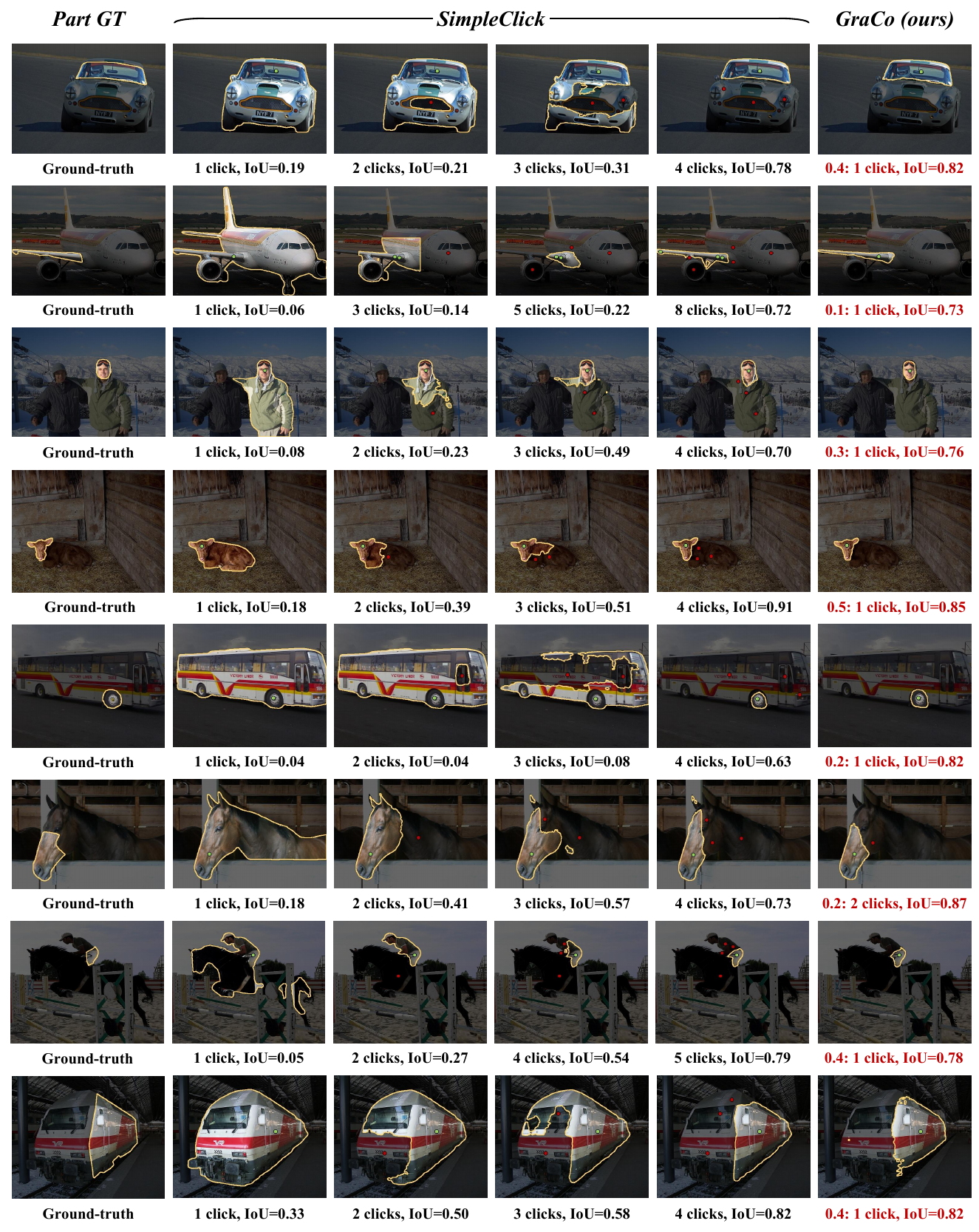}
    \caption{
        \textbf{More visualization examples of interactive segmentation on part GT using SimpleClick}~\cite{liu2023simpleclick} \textbf{and our GraCo.}
        The proposed  method satisfies the user's requirements with just one or two clicks.
    }
    \label{fig:part_vis_1}
\end{figure*}

\begin{figure*}[h]
    \centering
    \includegraphics[width=0.96\textwidth]{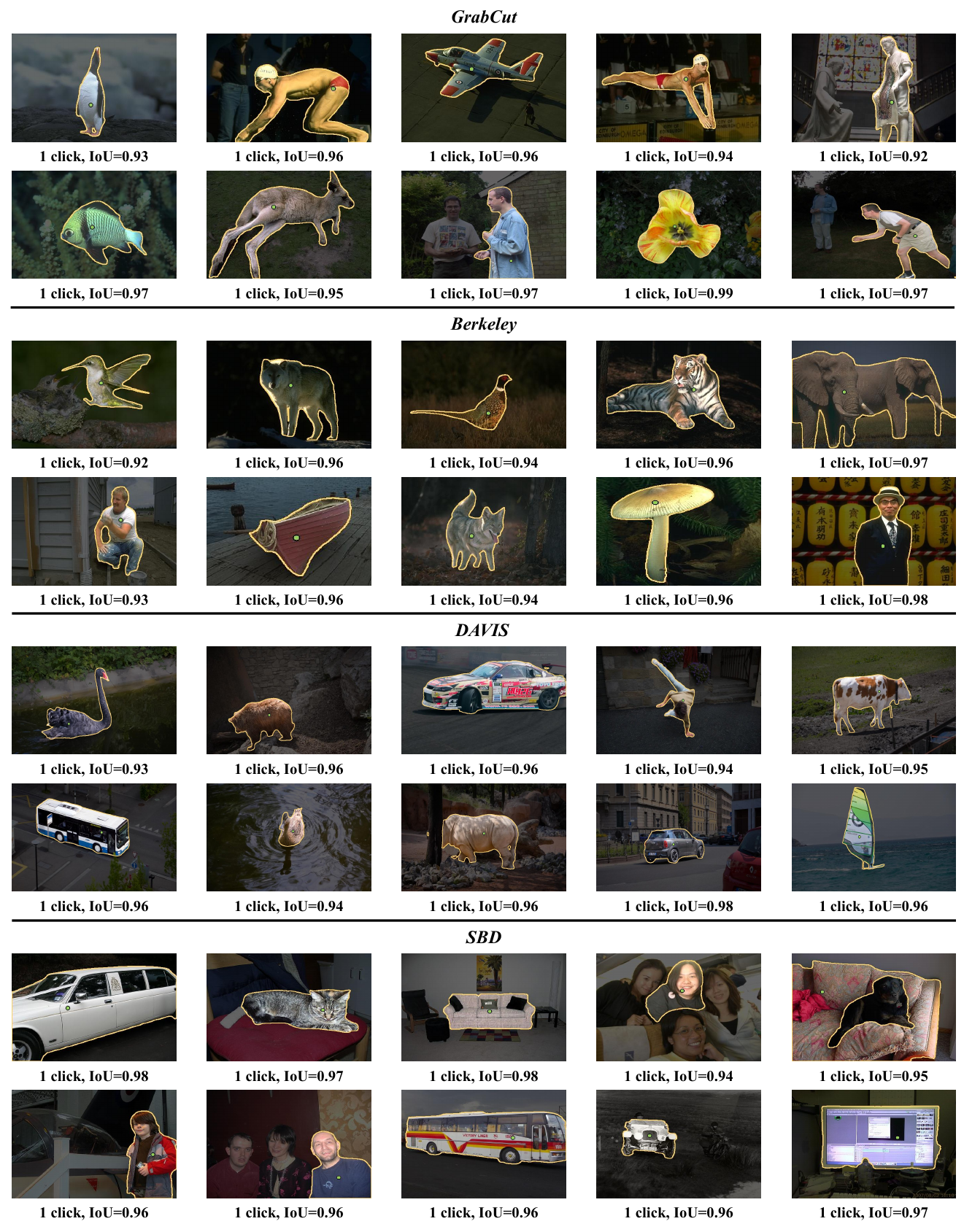}
    \caption{
        \textbf{Visualization on four object-level benchmarks.}
        Note that the input granularity of GraCo is fixed to 1.0.
    }
    \label{fig:vis_2}
\end{figure*}

\subsection{More Ablations}
\label{subsec:add ablation}
\noindent \textbf{Proposal Sampling.} 
We also conduct an ablation study on the proposal sampling. We compare the performance of uniform sampling to inverse-proportional sampling with identical mask proposals~(\textit{cf.}~\Cref{tab:sampling}).
The results show that the inverse-proportional sampling method achieves a superior performance on all benchmarks, which indicates that the method enables the IS model to learn uniformly from any-granularity proposals in GCL.
\smallskip

\noindent \textbf{LoRA.}
We supplement the ablation study on LoRA, as shown in~\Cref{tab:lora}. We employ identical AGG-generated mask proposals to train our GraCo equipped with ViT-B as backbone.
We set the LoRA rank as {4, 8, 16, 32}, respectively, and evaluate the performance on both levels of benchmarks.
Based on the results, we conclude that the performance of GraCo is not sensitive to the LoRA rank.
\smallskip

\noindent \textbf{Granularity Definition.}
We evaluate the performance of the two definitions on part-level benchmarks, which indicates that employing only scale granularity leads to a slight decrease~(\textit{cf.}~\Cref{tab:granularity}).
This demonstrates the necessity of the two types of granularity for definition.
\smallskip

\section{Dataset Description}
\label{sec:dataset}
We evaluate both object-level and part-level benchmarks to demonstrate the performance of the IS model in multi-granularity scenarios. The details of these datasets are described as follows.
\begin{itemize}
\item \textbf{GrabCut}~\cite{rother2004grabcut}. The dataset contains 50 images, each containing a single instance.
\item \textbf{Berkeley}~\cite{mcguinness2010comparative}. The dataset contains 96 images with 100 instances and some of them are more challenging for segmentation.
\item \textbf{SBD}~\cite{hariharan2011semantic}. The dataset contains 2,857 images with 6,671 challenging instances for evaluation and not be used for training.
\item \textbf{DAVIS}~\cite{perazzi2016benchmark}. The dataset contains 50 high-quality videos and we use 345 frames for evaluation.
\item \textbf{PascalPart}~\cite{chen2014detect}. The dataset provides part annotations of 20 Pascal VOC~\cite{everingham2010pascal} classes, a total of 193 part categories. As PascalPart contains a large number of parts, we randomly select 5 out of 16 classes~(excluding boat, chair, dining table, and sofa which do not have part annotations) to reduce the computational cost of conducting interactive simulations during evaluation. The selected classes are train, bicycle, cow, aeroplane, and bus in experiments.
\item \textbf{PartImageNet}~\cite{he2022partimagenet}. The dataset groups 158 classes from ImageNet~\cite{russakovsky2015imagenet} into 11 super-categories and provides a total of 40 part categories, which is a large, high-quality dataset for part segmentation, offering part-level annotations on a broad range of classes, including non-rigid, articulated objects. We use the validation set of PartImageNet to evaluate the performance of IS model at the part-level, which includes 1206 images and 5626 parts.
\item \textbf{SA-1B}~\cite{kirillov2023segment}. The dataset consists of 11M high-resolution~(3300×4950 pixels on average), diverse, and licensed images and 1.1B high-quality segmentation masks. To alleviate storage pressure, released images are downsampled and their shortest side is set to 1500 pixels. We use the first 1000 images to evaluate the performance of different methods.
\end{itemize}

\smallskip

\section{Additional Qualitative Results}
We supplement more examples to demonstrate the granularity controllability and excellent segmentation performance of our GraCo in multi-granularity scenarios, \textit{cf.}~\Cref{fig:part_vis_1}.
For complex scenarios, our GraCo allows the user to select the appropriate granularity to generate the required mask. Furthermore, our GraCo facilitates precise control over the expansion of segmentation masks through multiple positive clicks by applying a small granularity.
This advantage effectively overcomes the limitations of current object-level IS methods~(\eg, SimpleClick~\cite{liu2023simpleclick}) when dealing with tiny or detached components.
We also demonstrate the qualitative results of the proposed GraCo on four object-level benchmarks with a fixed input granularity of 1.0, \textit{cf.}~\Cref{fig:vis_2}.
Our GraCo achieves impressive qualitative results.

\end{document}